\documentclass[10pt,a4paper]{article}

\usepackage[top=2.0cm, bottom=2.0cm, left=2.0cm, right=2.0cm]{geometry}
\usepackage[utf8]{inputenc}
\usepackage[T1]{fontenc}
\usepackage[protrusion=true,expansion=false]{microtype}  
\usepackage{amsmath,amssymb,amsfonts,mathtools}
\usepackage{amsthm}
\IfFileExists{newtxtext.sty}{%
  \usepackage{newtxtext,newtxmath}  
}{}
\newtheorem{proposition}{Proposition}

\usepackage{booktabs}
\IfFileExists{multirow.sty}{\usepackage{multirow}}{}
\providecommand{\multirow}[3]{#3}
\usepackage{tabularx}
\usepackage{array}
\usepackage{colortbl}
\IfFileExists{makecell.sty}{\usepackage{makecell}}{}
\providecommand{\makecell}[2][]{\begin{tabular}{@{}c@{}}#2\end{tabular}}

\usepackage{graphicx}
\usepackage{subcaption}
\usepackage[font={footnotesize,it},labelfont={bf,up},labelsep=period,
            justification=justified,singlelinecheck=false,
            margin=0.5cm]{caption}
\usepackage{float}
\IfFileExists{placeins.sty}{%
  \usepackage{placeins}            
}{%
  \providecommand{\FloatBarrier}{}
}
\IfFileExists{needspace.sty}{%
  \usepackage{needspace}
}{%
  \newcommand{\Needspace}[1]{%
    \par\begingroup
    \dimen0=##1\relax
    \ifdim\pagegoal<\maxdimen
      \ifdim\dimexpr\pagegoal-\pagetotal\relax<\dimen0
        \break
      \fi
    \fi
    \endgroup}
}
\IfFileExists{standalone.sty}{\usepackage{standalone}}{}

\usepackage{array}
\usepackage[authoryear,round]{natbib}
\usepackage[dvipsnames]{xcolor}
\definecolor{navy}{HTML}{1b2a4a}
\definecolor{teal}{HTML}{1a7a6d}
\definecolor{dteal}{HTML}{0d4f47}
\definecolor{lgrey}{HTML}{f0f2f5}
\definecolor{mgrey}{HTML}{b0b8c4}
\definecolor{rmcol}{HTML}{9ca3af}
\definecolor{faccol}{HTML}{2d9d8f}
\definecolor{distcol}{HTML}{3b82c4}
\definecolor{retcol}{HTML}{e06060}
\definecolor{shadowcol}{HTML}{d0d4da}
\IfFileExists{enumitem.sty}{%
  \usepackage{enumitem}
  \setlist{noitemsep,topsep=2pt,parsep=0pt,partopsep=0pt}
}{%
  \let\origitemize\itemize
  \let\endorigitemize\enditemize
  \renewenvironment{itemize}[1][]{\origitemize}{\endorigitemize}
  \let\origenumerate\enumerate
  \let\endorigenumerate\endenumerate
  \renewenvironment{enumerate}[1][]{\origenumerate}{\endorigenumerate}
}
\usepackage{fancyhdr}
\IfFileExists{abstract.sty}{%
  \usepackage{abstract}
}{%
  
  \newlength{\absleftindent}
  \newlength{\absrightindent}
}
\IfFileExists{titlesec.sty}{\usepackage{titlesec}}{}

\PassOptionsToPackage{hyphens}{url}
\usepackage{url}

\usepackage{hyperref}
\hypersetup{
  colorlinks  = true,
  linkcolor   = NavyBlue,
  citecolor   = OliveGreen,
  urlcolor    = BrickRed,
  pdfauthor   = {Reza Barati, Qinmin Vivian Hu},
  pdftitle    = {gym-invmgmt: An Open Benchmarking Framework for Inventory Management Methods},
  pdfsubject  = {Inventory Management, Reinforcement Learning, Operations Research},
  pdfkeywords = {RL, benchmark, supply chain, GNN, stochastic programming}
}
\usepackage{bookmark}
\IfFileExists{cleveref.sty}{%
  \usepackage[capitalize]{cleveref}
  \crefname{figure}{Fig.}{Figs.}
  \Crefname{figure}{Fig.}{Figs.}
  \crefname{table}{Table}{Tables}
  \Crefname{table}{Table}{Tables}
  \crefname{section}{Section}{Sections}
  \Crefname{section}{Section}{Sections}
}{%
  \providecommand{\cref}[1]{\ref{##1}}
  \providecommand{\Cref}[1]{\ref{##1}}

}

\IfFileExists{titlesec.sty}{%
  \titleformat{\section}{\large\bfseries}{\thesection}{1em}{}
  \titleformat{\subsection}{\normalsize\bfseries}{\thesubsection}{1em}{}
  \titlespacing*{\section}{0pt}{1.2ex plus 0.3ex minus .2ex}{0.6ex plus .2ex}
  \titlespacing*{\subsection}{0pt}{0.8ex plus 0.2ex minus .2ex}{0.4ex plus .2ex}
}{}

\usepackage{pifont}
\newcommand{\cmark}{\ding{51}}
\newcommand{\xmark}{\ding{55}}

\definecolor{lightgray}{gray}{0.93}
\definecolor{tableheader}{HTML}{2C3E50}
\definecolor{tableaccent}{HTML}{E8F4FD}
\definecolor{headerblue}{HTML}{1A5276}

\title{\textbf{gym-invmgmt: An Open Benchmarking\\[4pt]
       Framework for Inventory Management Methods}}

\author{%
  Reza Barati\thanks{Corresponding author:
  \texttt{r2barati@torontomu.ca}}
  \quad Qinmin Vivian Hu\\[4pt]
  \small Department of Computer Science\\[-1pt]
  \small Toronto Metropolitan University, Toronto, ON, Canada
}

\date{%
  \small\today\\[4pt]
  \small\textcolor{gray}{Preprint.}%
}

\begin{document}

\maketitle
\thispagestyle{fancy}

\begin{abstract}
\noindent
Inventory-policy comparisons are often difficult to interpret because
performance depends on the evaluation contract as much as on the policy
itself. Differences in topology, demand regime, information access,
feasibility constraints, shortage treatment, and Key Performance
Indicator~(KPI) definitions can change method rankings. We present
\textbf{\texttt{gym-invmgmt}}, a
Gymnasium-compatible extension of the OR-Gym inventory-management lineage for auditable
cross-paradigm evaluation. The benchmark evaluates optimization,
heuristic, and learned controllers under a shared CoreEnv transition,
reward, action-bound, and KPI contract, while varying stress conditions
through a 22-scenario core grid plus four supplemental MARL-mode rows.
Within these released scenarios,
informed stochastic programming provides the strongest non-oracle reference,
reflecting the value of scenario hedging under forecast access, but at
substantially higher online computational cost. Among
learned controllers, the Proximal Policy Optimization Transformer variant
(PPO-Transformer) achieves the strongest learned-policy quality at fast
inference, while Residual Reinforcement Learning~(Residual~RL) provides
competitive hybrid performance. The graph neural network variant
(PPO-GNN) is highly competitive on the default divergent topology but less
robust on the serial topology. Imitation learning performs well in
stationary regimes but degrades under demand shift, and the bounded Large
Language Model~(LLM) policy-parameter baseline is best interpreted as a
diagnostic controller rather than an autonomous inventory optimizer.
Overall, the benchmark identifies scenario-conditioned leaders
while showing that performance depends jointly on information access, demand
shift, topology, and policy representation.
\end{abstract}


\vspace{0.3em}
\noindent\rule{\textwidth}{0.3pt}
\vspace{0.5em}

\section{Introduction}\label{sec:intro}

Multi-echelon inventory control is difficult not only because the decision
problem is stochastic and sequential~\citep{clark1960,zipkin2000}, but because
policy performance is highly context-dependent. A replenishment rule that
performs well under stationary demand, short lead times, and backlog
fulfillment may fail under demand shocks, lost sales, or topology changes;
similarly, a method that appears strong in one simulator may benefit from
visibility, action-bound, demand-generation, or cost-accounting assumptions
absent in another. Comparing policies therefore requires a shared experimental
contract, not only a reported profit or service level.

Recent OR/RL benchmark efforts have moved toward such contracts:
\citet{balaji2020} introduced reusable RL benchmarks for online stochastic
optimization; \citet{hubbs2020} introduced OR-Gym, including serial
multi-echelon inventory; and \citet{perez2021} extended the OR-Gym inventory
setting toward tree/network topologies while comparing deterministic linear
programming, Multi-Stage Stochastic Programming~(MSSP), and reinforcement
learning under shared stochastic simulation mechanics. These works establish
the lineage of our benchmark rather than serving as a contrast to it.

The remaining limitation is the narrowness of the comparison contract. Prior
environments typically expose only part of the experimental space: stationary
synthetic demand, flat observations, a fixed topology family, centralized
OR/RL comparison, or decentralized multi-agent coordination. MARL benchmarks
such as MABIM~\citep{liu2023mabim} and recent multi-agent supply-chain
studies~\citep{kotecha2025,quan2024} expand the decentralized side, but their
scientific focus differs from controlled centralized cross-paradigm
comparison.

Building on this lineage, we present \texttt{gym-invmgmt},\footnote{Benchmark framework, agents, and evaluation scripts:
\url{https://github.com/r2barati/gym-invmgmt-paper}.  A standalone Gymnasium
environment library covering Newsvendor, Multi-Echelon, and Network
inventory problems is available separately at
\url{https://github.com/r2barati/gym-invmgmt}.} a
Gymnasium-compatible extension for quality--speed--robustness comparison
under curated stress scenarios and explicit blind/informed information
protocols. Optimization baselines use the same scenario contract, while
learned policies are trained as generalist checkpoints and evaluated without
scenario-specific retraining. The main study keeps centralized control as the
experimental lens, isolating inventory optimization from coordination effects
while retaining a per-node CTDE wrapper for decentralized diagnostics.

\paragraph{Contributions.}
\begin{enumerate}[leftmargin=*,itemsep=0pt,topsep=1pt]
  \item \textbf{Benchmark contract.}
        We extend the OR-Gym inventory line with configurable DAG topologies,
        stochastic and empirical demand, backlog/lost-sales modes, endogenous
        goodwill, and vector, graph, and sequence observations.

  \item \textbf{Released scenario matrix.}
        We release 26 stress scenarios (22~core plus four MARL-mode rows)
        for 29 registered non-LLM configurations and one bounded
        LLM-policy-parameter baseline under shared action, demand, KPI, and
        information protocols.

  \item \textbf{Generalist learned-policy evaluation.}
        PPO, SAC, Transformer, GNN, Residual, and imitation-learning
        checkpoints are evaluated without per-scenario retraining; a separate
        zero-shot graph-transfer stress test diagnoses cross-topology
        action-decoding limits.

  \item \textbf{Cross-paradigm findings.}
        MSSP-I is the strongest non-oracle reference, information access most
        helps stochastic programming and variance-sensitive heuristics, and
        learned/LLM/transfer tests expose covariate-shift, service-level, and
        action-decoding limits.
\end{enumerate}

\section{Background and Related Work}\label{sec:background}

Prior inventory-control work offers several useful but differently
instrumented views of the same decision problem. The classical OR view treats
inventory as a sequential decision problem, formalized by dynamic programming
and Markov decision processes~\citep{bellman1957,puterman1994}.
Mathematical-programming work emphasizes planning, recourse, and safety-stock
placement in networks~\citep{graves2000}. Replenishment-policy theory adds
interpretable controls: Scarf's min--max model is an early robust newsvendor
formulation~\citep{scarf1958}, Zipkin systematizes inventory-policy structure
~\citep{zipkin2000}, and Porteus develops stochastic inventory-theory
foundations~\citep{porteus2002}. Simulation-based learning changes the lens:
Giannoccaro and Pontrandolfo learn supply-chain ordering decisions through
interaction~\citep{giannoccaro2002}, and Kemmer et al.\ extend this framing in
later inventory-control experiments~\citep{kemmer2018}. Subsequent deep RL
studies address multi-echelon demand uncertainty~\citep{gao2020}, spare-parts
inventory~\citep{wang2021}, and multi-product, lead-time-constrained networks
~\citep{meisheri2022}. Data-driven inventory optimization further shows how
machine-learning decision rules can move beyond purely parametric newsvendor
assumptions~\citep{ban2019}.

\begin{figure}[!htbp]
  \centering
  \includegraphics[width=0.85\linewidth]{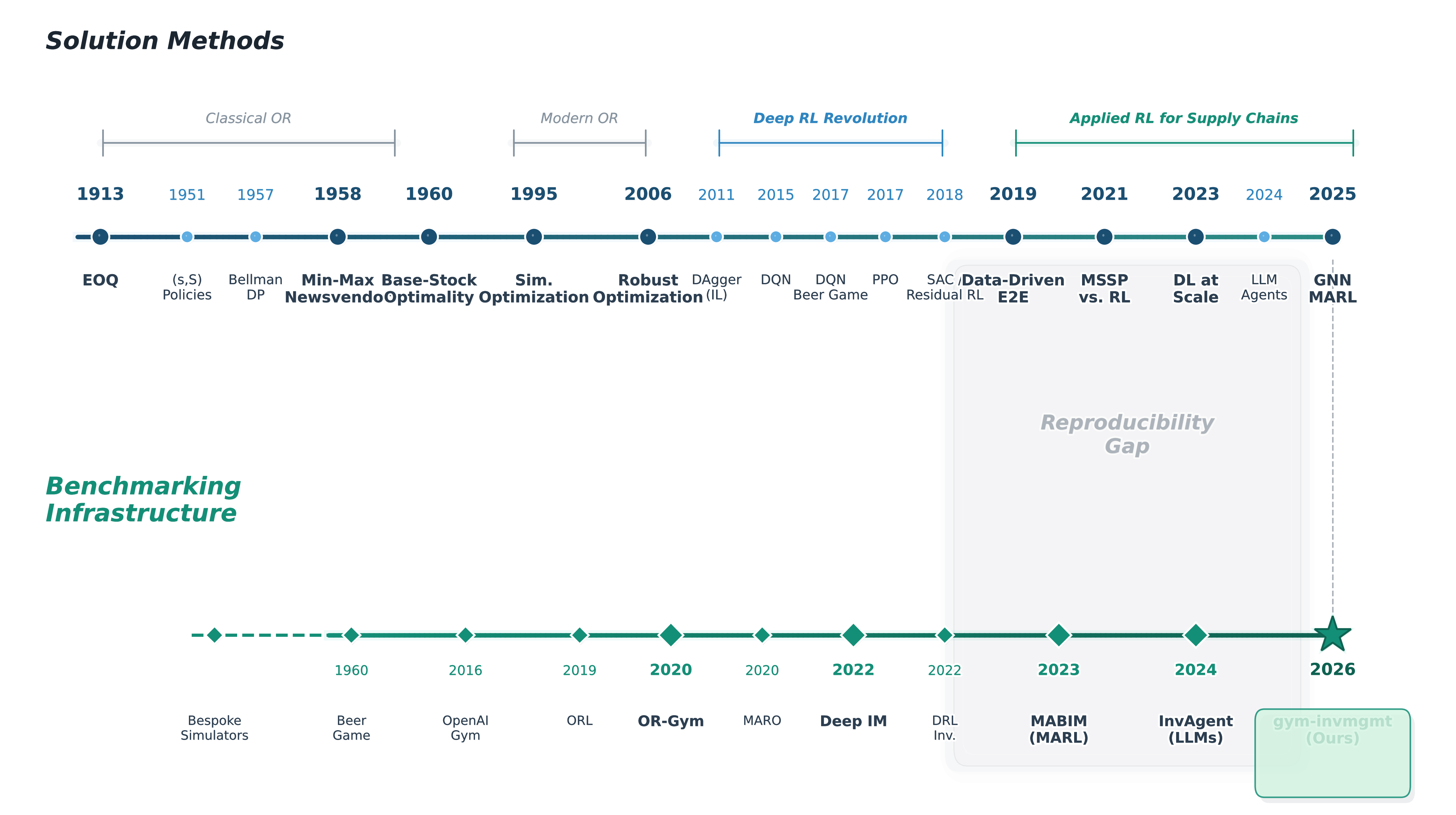}
  \caption[Representative timeline of inventory solution methods and benchmarking infrastructure.]{%
    Inventory-method and benchmark-infrastructure timeline, from
    EOQ~\citep{harris1913} to Gym-style OR benchmarks~\citep{brockman2016}.}
  \label{fig:timeline}
\end{figure}

Complementary inventory-learning studies expand important parts of this
landscape: beer-game RL~\citep{oroojlooyjadid2022}, deep RL versus base-stock
policies~\citep{gijsbrechts2022}, decentralized MARL benchmarks
~\citep{liu2023mabim,leluc2022}, graph-based MARL~\citep{kotecha2025}, and
LLM-based inventory agents~\citep{quan2024}. Their focus, however, is often
decentralized coordination, a specific topology, a single method family, or
language-agent reasoning rather than centralized cross-paradigm evaluation
under one transition, reward, action-bound, information-access, and KPI
contract.

Taken together, this breadth makes numerical comparison fragile. Review and
roadmap papers
emphasize that DRL inventory performance depends on modeling choices, tuning
burden, and evaluation design~\citep{boute2022}, and that supply-chain RL
studies vary widely in formulation and assessment protocol~\citep{rolf2023}.
Adjacent inventory-control studies show the same issue in other forms:
centralization can change replenishment behavior~\citep{chen2005}, while
inventory-routing formulations often embed problem-specific state, routing,
and service conventions~\citep{charaf2024}. Thus, apparent performance
differences may reflect the evaluation contract as much as the policy class
itself.

\Cref{tab:benchmark_comparison} therefore maps representative studies along
the evaluation axes needed for cross-paradigm comparison: topology, demand,
action constraints, data scope, lead times, horizon protocol, observations,
endogenous feedback, and evaluated algorithms. It is not a one-to-one
performance comparison; it shows which parts of the experimental contract each
line of work makes explicit.

\begin{table}[H]
\centering
\caption[Comparison of \texttt{gym-invmgmt} to existing benchmarks and studies.]{%
  Comparison of \texttt{gym-invmgmt} to representative inventory benchmarks
  and studies.}
\label{tab:benchmark_comparison}
\renewcommand{\arraystretch}{0.88}
\setlength{\tabcolsep}{2.2pt}
\resizebox{\textwidth}{!}{%
\scriptsize
\begin{tabular}{@{} >{\raggedright\arraybackslash}p{2.5cm} c c c c c c c c c c p{3.2cm} @{}}
\toprule
\makecell{\textbf{Benchmark}\\[-1pt]\textbf{/ Study}}
  & \textbf{Year}
  & \makecell{\textbf{Network}\\[-1pt]\textbf{Topology}}
  & \makecell{\textbf{Agent}\\[-1pt]\textbf{Formul.}}
  & \makecell{\textbf{Demand}\\[-1pt]\textbf{Model}}
  & \makecell{\textbf{Capacity/}\\[-1pt]\textbf{Action}\\[-1pt]\textbf{Constr.}}
  & \makecell{\textbf{SKU \&}\\[-1pt]\textbf{Data}}
  & \makecell{\textbf{Lead Times}\\[-1pt]\textbf{\& Shortages}}
  & \makecell{\textbf{Horizon}\\[-1pt]\textbf{Protocol}}
  & \makecell{\textbf{Obs.}\\[-1pt]\textbf{Spaces}}
  & \makecell{\textbf{Endog.}\\[-1pt]\textbf{Demand}\\[-1pt]\textbf{Feedback}}
  & \centering\arraybackslash\textbf{Evaluated Algorithms} \\
\midrule
\rowcolor{lightgray}
ORL~\citep{balaji2020}
  & 2020
  & Single-Node & Central. & Station.\ Pois.
  & Unconstr. & 1-SKU, Synth.
  & Fixed VLT, LS & Finite online & Flat Vec. & \xmark
  & PPO, APE-X DQN, heuristic/MIP baselines \\
OR-Gym~\citep{hubbs2020}
  & 2020
  & Multi-Ech. & Central. & Station.\ Stoch.
  & Prod./inv.\ cap. & 1-SKU, Synth.
  & Fixed LT, BL/LS & Finite SH & Flat Vec. & \xmark
  & PPO, heuristics, MIP/SH baselines \\
\rowcolor{lightgray}
Perez inventory~\citep{perez2021}
  & 2021
  & Multi-Ech. & Central. & Station.\ Stoch.
  & Prod./inv.\ cap. & 1-SKU, Synth.
  & Het.\ LT, BL/LS & Finite RH/SH & Flat Vec. & \xmark
  & DLP/MSSP (RH/SH), PPO, Oracle \\
IM~Sim.~\citep{sridhar2021}
  & 2021
  & Single-Store & Sim./Opt. & Case data
  & Svc.\ constr. & 1 item, Real
  & Stoch.\ LT, LS & Finite DES & Sim.\ state & \xmark
  & Arena sim., OptQuest \\
\rowcolor{lightgray}
MARLIM~\citep{leluc2022}
  & 2022
  & Single-Ech. & Decentr. & Station.\ ZIP
  & Storage cap. & 50 items, Real-fit
  & Stoch.\ LT, BL/overflow & Finite episodic & Flat Vec. & \xmark
  & PPO-D/C, IPPO-C, MinMax, Oracle \\
Deep IM~\citep{madeka2022}
  & 2022
  & Single-Ech. & Central. & Hist./Non-Stat.
  & Unconstr. & Multi-SKU, Real
  & Stoch.\ VLT, LS & Historical rollout & Hist.\ Vec. & \xmark
  & DirectBackprop, A3C, SAC, ARS, Newsvendor \\
\rowcolor{lightgray}
DQN~Beer~\citep{oroojlooyjadid2022}
  & 2022
  & Multi-Ech. & Decentr. & Station./Step/Real
  & Unconstr. & 1-SKU, Synth./Real
  & Det.\ LT, BL & Finite episodic & \makecell{Local Hist.\\[-2pt]Vec.} & \xmark
  & SRDQN, BS co-players \\
DRL~Inv.~\citep{gijsbrechts2022}
  & 2022
  & Multi-Ech. & Central. & Station.
  & Feas.\ alloc. & 1-SKU, Synth.
  & Fixed LT, BL/LS & Finite episodic & Flat Vec. & \xmark
  & A3C, base-stock \\
\rowcolor{lightgray}
MengQi~E2E~\citep{qi2023}
  & 2023
  & Single-Ech. & Central. & Hist./Non-Stat.
  & Unconstr. & Multi-SKU, Real
  & Stoch.\ VLT, LS & Finite window & Feature Vec. & \xmark
  & E2E ML, PTO, heuristics \\
MABIM~\citep{liu2023mabim}
  & 2023
  & Multi-Ech. & Decentr. & Station./Non-Stat.
  & Warehouse cap. & 2000+ SKU, Real
  & Var.\ LT, BL/overflow & Finite episodic & SKU/WH Vec. & \xmark
  & IPPO, QTRAN, BS, $(s,S)$, hybrid BS+MARL \\
\rowcolor{lightgray}
InvAgent~\citep{quan2024}
  & 2024
  & Multi-Ech. & Decentr. & Multi-Regime
  & Prod.\ cap. & 1-SKU, Synth.
  & Fixed LT, BL & Finite episodic & Text prompt & \xmark
  & GPT-4/4o LLMs, IPPO/MAPPO, BS/tracking \\
GNN~MARL~\citep{kotecha2025}
  & 2025
  & Multi-Ech. & Decentr./CTDE & Stoch./Shock
  & Prod.\ cap. & 1-SKU, Synth.
  & Stoch.\ LT, BL & Finite episodic & \makecell{Network\\[-2pt]Graphs} & \xmark
  & GNN-MAPPO, IPPO/MAPPO, heuristics \\
\midrule
\makecell[tl]{\textbf{\texttt{gym-invmgmt}}\\[-2pt]\textbf{(Ours)}}
  & \makecell[tc]{\textbf{2026}}
  & \makecell[tc]{\textbf{Multi-Ech.}} & \makecell[tc]{\textbf{Central.}\\[-2pt]\textbf{(+MARL)}}
  & \makecell[tc]{\textbf{Stat.,}\\[-2pt]\textbf{Non-Stat.,}\\[-2pt]\textbf{External}}
  & \makecell[tc]{\textbf{Prod./action}\\[-2pt]\textbf{cap.}}
  & \makecell[tc]{\textbf{1-SKU,}\\[-2pt]\textbf{Synth./}\\[-2pt]\textbf{M5 traces}}
  & \makecell[tc]{\textbf{Fixed LT,}\\[-2pt]\textbf{BL/LS}}
  & \makecell[tc]{\textbf{Finite}\\[-2pt]\textbf{episodic/RH}}
  & \makecell[tc]{\textbf{Graph, Seq.,}\\[-2pt]\textbf{Vec., Text}}
  & \makecell[tc]{\textbf{\cmark}\\[-2pt]\textbf{Goodwill}\\[-2pt]\textbf{dyn.}}
  & \textbf{Oracle, DLP/MSSP, heuristics, PPO/SAC, GNN/Transformer/ST-PPO, Residual RL, DAgger/GNN-IL, LLM-Policy-C diagnostics} \\
\bottomrule
\end{tabular}}
\vspace{2pt}
\begin{minipage}{0.98\textwidth}
\scriptsize
\emph{Note.} Rows summarize each work's primary experimental setting or
released benchmark contract, not every variant studied. Endogenous demand
feedback means realized service changes future demand; static stockout
penalties, service constraints, and historical arrivals alone are not counted.
The \texttt{gym-invmgmt} row reports the main centralized benchmark plus
released diagnostic wrappers/agents. The Horizon Protocol column reports
finite episodic/window evaluation and whether optimization baselines use
rolling or shrinking replanning; RH/SH denote rolling/shrinking optimization
horizons, while Oracle denotes a full-horizon non-causal reference. Abbrev.:
Ech.=echelon; Formul.=formulation; Constr.=constraints; Obs.=observation;
Endog.=endogenous; Central./Decentr.=
centralized/decentralized; Stat./Non-Stat.=stationary/non-stationary; Det.=
deterministic; Stoch.=stochastic; Pois.=Poisson; ZIP=zero-inflated Poisson;
SKU=stock-keeping unit; Synth.=synthetic; LT/VLT=lead/vendor lead time;
Het./Var.=heterogeneous/variable; BL/LS=backlog/lost sales; Unconstr.=
unconstrained; Prod./inv.=production/inventory; cap.=capacity; Svc.=
service-level; Feas.=feasibility-enforced; Hist.=historical; Vec./Seq.=
vector/sequence; DES=discrete-event simulation; Sim.=simulation; WH=warehouse;
BS=base-stock; LLM=large language model; MARL=multi-agent reinforcement
learning; MIP=mixed-integer programming; PTO=predict-then-optimize; RH/SH=
rolling/shrinking horizon.
\end{minipage}
\end{table}

\FloatBarrier

\section{Benchmark Framework and Environment Contract}\label{sec:framework}

The framework is built around a single finite-horizon environment,
\texttt{CoreEnv}, defined on a directed supply-chain graph. Nodes represent
production, storage, retail, or market locations; edges encode replenishment
flows, retail demand links, lead times, costs, capacity/yield limits, and
pipeline inventory. At each period, a controller submits a bounded reorder
vector over the replenishment edges. The same transition code then applies
feasibility constraints, material-balance updates, reward accounting, and KPI
extraction regardless of whether the controller is an optimizer, heuristic, or
learned policy.

Scenario definitions vary the experimental conditions around this shared core:
topology, backlog versus lost-sales treatment, information access, demand
process, external demand traces, and endogenous goodwill feedback. This
separation lets the benchmark compare policy classes under matched physical
mechanics rather than under method-specific state, reward, or accounting
conventions.

\begin{figure}[!htbp]
  \centering
  \includegraphics[width=0.95\linewidth]{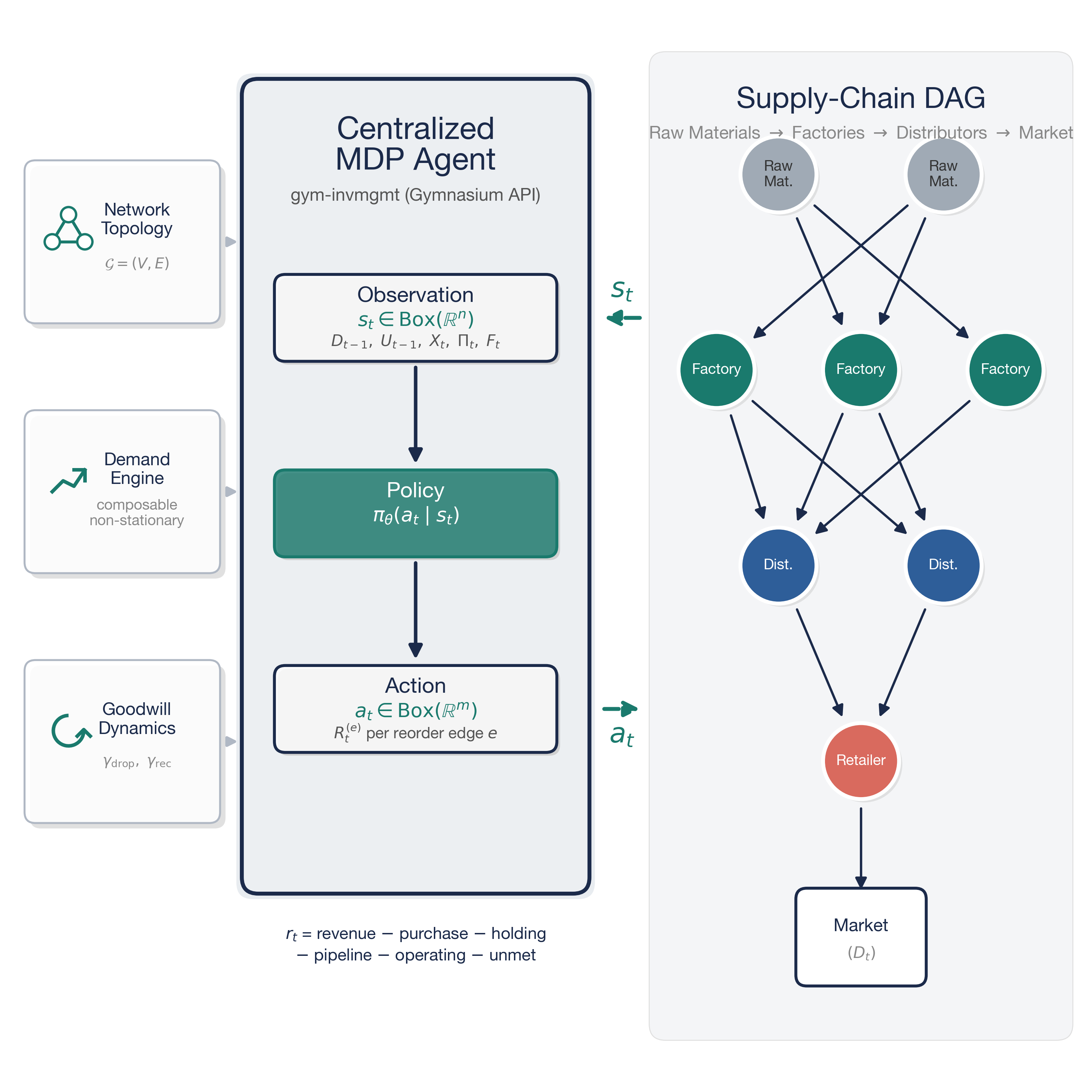}
  \caption[Schematic architecture of the \texttt{gym-invmgmt} framework.]{%
    Schematic architecture of the \texttt{gym-invmgmt} framework.
    Topology, demand, and goodwill modules feed a centralized
    \texttt{CoreEnv} Gymnasium~\citep{towers2023} MDP~\citep{puterman1994};
    agents observe inventory/pipeline state and reorder on active links.
    Right: default divergent topology.}
  \label{fig:framework}
\end{figure}

\begin{figure}[!htbp]
  \centering
  \begin{subfigure}[t]{0.48\linewidth}
    \centering
    \includegraphics[width=\linewidth]{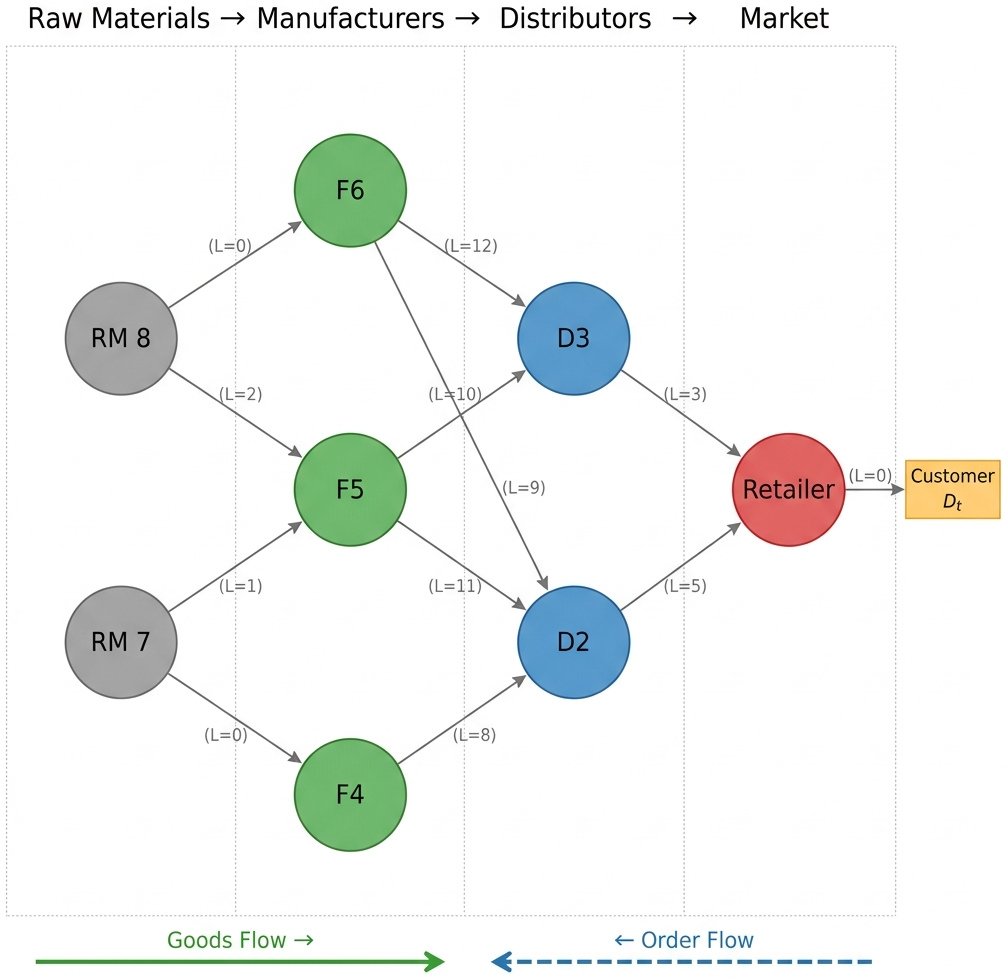}
    \caption{Default multi-echelon network topology
      (9~nodes, 12~total edges: 11~reorder links plus one
      retail-demand link, 4~echelons). Edge labels denote lead
      times~($L$). The legend indicates downstream goods flow and
      upstream order flow.}
    \label{fig:topologies}
  \end{subfigure}
  \hfill
  \begin{subfigure}[t]{0.48\linewidth}
    \centering
    \includegraphics[width=\linewidth]{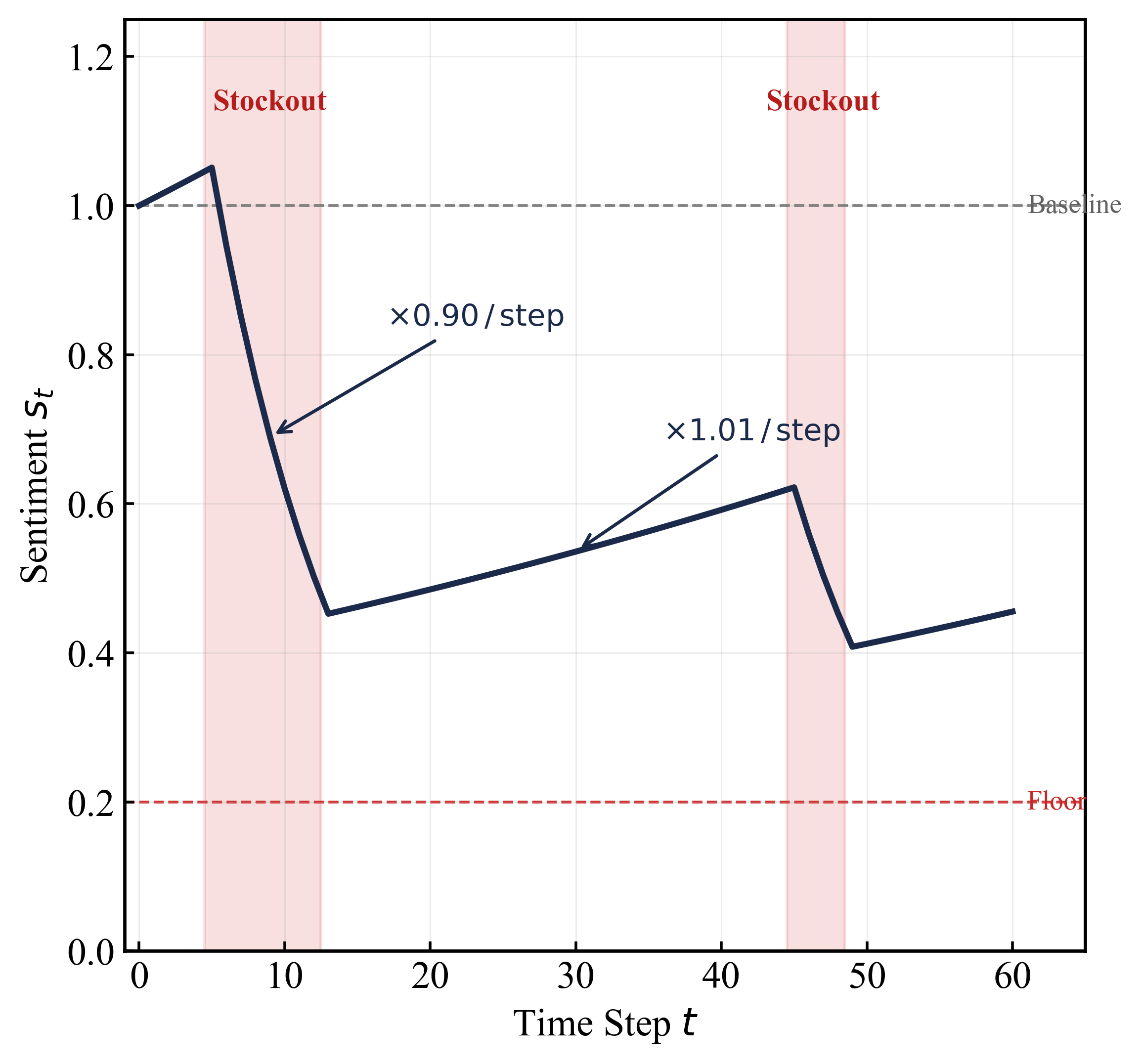}
    \caption{Illustrative asymmetric sentiment dynamics under endogenous
      customer goodwill. Stockout episodes (shaded) trigger
      abrupt decay ($\times 0.90$/step); recovery proceeds
      slowly ($\times 1.01$/step). Floor at $s_{\min} = 0.2$.}
    \label{fig:goodwill_dynamics}
  \end{subfigure}
  \caption[Supply chain topology and endogenous goodwill dynamics.]{%
    \emph{(a)}~Default network topology used in the benchmark.
    Raw material suppliers feed three capacity-constrained factories,
    which ship through two distributors to a single retailer facing
    stochastic demand~$D_t$.
    \emph{(b)}~Endogenous customer goodwill dynamics. The asymmetry
    between rapid sentiment decay and slow recovery rewards policies
    that protect service levels before stockouts compound.}
  \label{fig:topology_and_goodwill}
\end{figure}

\subsection{Benchmark Scope and Information Protocol}\label{sec:information_protocol}

The main benchmark adopts centralized control as a methodological lens.
At each period, one controller issues replenishment requests for all
reorder edges in the supply-chain graph. This choice is not a claim that
real supply chains are always centrally controlled; rather, it removes
decentralized coordination as a confound when comparing optimization
models, heuristics, imitation learning, and neural policies under the
same transition and reward contract. The per-node multi-agent wrapper is
retained for diagnostic and future decentralized experiments, but the
headline results use the centralized setting to isolate the
inventory-control problem itself.

Information access is reported separately from policy architecture.
\emph{Blind} agents act from realized demand history, pipeline orders,
and current inventory state. \emph{Informed} agents may use benchmark
demand-context features or the specified demand model, but still act
causally at each decision period. The Oracle is non-causal: in
exogenous-demand settings it uses future demand realizations as a
full-horizon reference, while in endogenous-goodwill settings it is
interpreted as a clairvoyant simulation-optimization benchmark rather
than a certified dynamic-programming optimum. This separation prevents
the evaluation from conflating three axes: coordination scope,
demand-model access, and architectural inductive bias.

The notation follows the inventory-management MDP lineage of
OR-Gym and \citet{hubbs2020,perez2021}, while the formal
finite-horizon MDP conventions follow standard stochastic dynamic
programming notation~\citep{puterman1994}. We restate the transition and
reward contract here because the benchmark extends that lineage to graph
topologies, multiple observation encodings, empirical demand traces, and
endogenous-goodwill stress tests. Formally,
\texttt{CoreEnv} defines a graph-augmented finite-horizon MDP:
\begin{equation}\label{eq:mdp}
  \mathcal{M} = \bigl(G,\; \mathcal{S},\; \mathcal{A},\;
                       \mathcal{P},\; r,\; \alpha\bigr)
\end{equation}
where $G=(V,E)$ is a directed supply-chain graph, $E_r\subseteq E$
denotes the reorder edges, $\mathcal{S}\subseteq\mathbb{R}^{d_s}$ is
the continuous state or wrapped observation space, and
$\mathcal{A}\subseteq\mathbb{R}^{|E_r|}$ is the bounded continuous
reorder space. The transition kernel
$\mathcal{P}(\cdot\mid s_t,a_t,G)$ combines stochastic or empirical
demand generation with deterministic material-balance, pipeline,
feasibility, and shortage updates. The reward function~$r$ records
period profit after revenue, procurement, operating, holding, shortage,
pipeline, and optional fixed-ordering costs, with discount factor
$\alpha$.

Including $G$ as a structural parameter makes topology part of the
benchmark contract rather than an implicit simulator detail. Observation
wrappers can expose this same physical state as flat vectors, graph
features, per-link features, temporal stacks, or text prompts, allowing
architecture comparisons without changing the underlying transition
mechanics. All benchmark episodes remain finite-horizon simulations:
learned policies act causally one period at a time, rolling-horizon OR
agents repeatedly solve capped lookahead problems from the current
state, and the Oracle provides the non-causal full-horizon reference
where appropriate.

\subsection{Graph Topologies and Observation--Action Interfaces}

The framework models supply chains as Directed Acyclic Graphs~(DAGs),
natively defining classical serial chains and the default
multi-echelon network visualized in \Cref{fig:topologies}. We
partition the edge set into reorder links $E_r$, which carry upstream
replenishment requests and downstream physical shipments subject to lead
times, and retail-demand links $E_d$, which connect retailer nodes to
external markets. Topologies are parameterized procedurally or through
YAML specifications, allowing researchers to define custom adjacency
structures, node capacities, edge lead times, and retail demand
attributes without changing the underlying transition code.

At each discrete time step~$t$, the base environment constructs the
causal observation vector
\[
  o_t^{\mathrm{base}}
    = [D_{t-1},\, U_{t-1},\, X_t,\, \Pi_t,\, F_t],
\]
where $D_{t-1}$ is the last realized retail demand, $U_{t-1}$ records
previous-period unfulfilled demand (standing backlog in backlog mode and
lost demand in lost-sales mode), $X_t$ is current on-hand inventory,
$\Pi_t$ is an arrival-indexed pipeline schedule over reorder links, and
$F_t$ contains time and goodwill/sentiment features returned by the demand
engine. The pipeline schedule is deliberately reconstructed from
physically filled orders~$R$, not raw requests~$a$, so agents observe
committed in-transit inventory rather than phantom inventory that was
requested but never shipped because of upstream inventory, yield, or
capacity constraints.

Observation wrappers expose this same simulator state through different
interfaces. Flat-vector agents consume the base and domain-feature
vectors directly; graph and per-link wrappers reshape the same state into
node, edge, or link features; temporal wrappers stack recent observations
for sequence policies; and text interfaces serialize selected state
variables for bounded LLM-policy diagnostics. The blind/informed protocol
from \Cref{sec:information_protocol} is applied through the demand-access
component of these interfaces: blind agents estimate demand context from realized
history, while informed agents may use benchmark demand-context features
or the specified demand model. Thus, architecture comparisons change the
observation interface without changing the underlying transition
mechanics.

Based on this observation, continuous replenishment actions~$a_t^{(e)}$
are issued for every reorder edge~$e \in E_r$. These actions represent
raw replenishment requests. The environment records them for auditability,
then converts them into physically filled orders~$R_t^{(e)}$ after
applying non-negativity, upstream inventory, yield, and capacity
constraints. Neural agents are typically trained through a standard
$[-1,1]$ rescaled action interface for stable continuous-control
optimization, but the shared benchmark contract evaluates all controllers
through the native non-negative reorder-action space.

The resulting state vector has dimensionality
$d_s = \mathcal{O}\bigl(|V| + |E_d| + |E_r|L_{\max} + d_F\bigr)$,
where $L_{\max}$ is the longest reorder-edge lead time and $d_F$ is the
demand-engine feature dimension. This reflects node-level inventory,
retail-demand and unfulfilled-demand features, and explicit pipeline
tracking per reorder edge and lag period. The representation remains
linear in the graph and pipeline size, while stochastic-programming
scenario trees such as MSSP grow rapidly with planning horizon, branching
factor, and network size~\citep{birge2011}.

\subsection{Transition Dynamics and Material Balance}\label{sec:dynamics}

To ensure exactness, the environment enforces deterministic material-balance
updates inside the stochastic transition kernel~$\mathcal{P}$, following
classical inventory conservation logic~\citep{zipkin2000}. Let
$\operatorname{pred}(j)$ and $\operatorname{succ}(j)$ denote the immediate
predecessors and successors of node~$j$ in~$G$. Within each period, the
simulator applies a fixed causal event sequence: (1) raw replenishment requests
are allocated into feasible filled orders; (2) lead-time deliveries and pipeline
stocks are updated; (3) goodwill is updated from the previous period's
unfulfilled demand; (4) current retail demand is realized and retail fulfillment
is recorded; and (5) period rewards are computed.

For each reorder edge~$e=(i,j)$, the feasible filled order is
\begin{equation}\label{eq:filled_order}
  R_t^{(e)} = \min\left\{\max(a_t^{(e)},0),\; \kappa_t^{(e)}\right\},
\end{equation}
where $\kappa_t^{(e)}$ is the link-specific feasible supply bound. External
raw-material links fill any non-negative request, distributor links are capped
by currently available upstream inventory, and factory links are capped by both
remaining production capacity and yield-adjusted input inventory. This
distinction between requested orders~$a_t$ and filled orders~$R_t$ is central to
the benchmark contract.

For each managed node~$j \in M$ (retailers, distributors, and factories), the
on-hand inventory transition to~$t{+}1$ is governed by:
\begin{equation}\label{eq:material_balance}
  X_{t+1}^{(j)} \;=\; X_t^{(j)}
    \;+\; \sum_{i \in \operatorname{pred}(j)} R_{t-L_{ij}}^{(i,j)}
    \;-\; \frac{1}{v_j}
           \sum_{k \in \operatorname{succ}(j)} S_t^{(j,k)}
\end{equation}
where $R_{\tau}^{(i,j)}=0$ for $\tau<0$,
$R_{t-L_{ij}}^{(i,j)}$ represents incoming deliveries from filled orders placed
$L_{ij}$ periods earlier, $S_t^{(j,k)}$ represents outgoing shipments fulfilled
to downstream successors, and $v_j \in (0,1]$ is the production yield factor
($v_j=1$ for non-production nodes). Shipments are recorded in output units; a
factory shipment of $S_t^{(j,k)}$ therefore consumes $S_t^{(j,k)}/v_j$ units of
input inventory. Simultaneously, the in-transit pipeline inventory~$Y$ for each
reorder edge~$e \in E_r$ updates as:
\begin{equation}\label{eq:pipeline}
  Y_{t+1}^{(e)} \;=\; Y_t^{(e)} \;-\; R_{t-L_e}^{(e)}
                  \;+\; R_t^{(e)} .
\end{equation}

Customer demand is then sampled on each retail-demand link. If backlog mode is
enabled, the effective demand $\widetilde D_t^{(j,k)}$ includes the previous
period's unmet demand; otherwise, unmet demand is treated as lost sales and is
not carried forward. Retail shipments satisfy
$S_t^{(j,k)}=\min\{\widetilde D_t^{(j,k)},X_{t+1}^{(j)}\}$, and the unfulfilled
gap is recorded as
$U_t^{(j,k)}=\widetilde D_t^{(j,k)}-S_t^{(j,k)}$. Thus $U_t$ is always tracked
for penalty, service, and goodwill diagnostics, but it only becomes part of
$\widetilde D_{t+1}$ in backlog configurations. These transition constraints
ensure that neural policies, heuristics, and mathematical programming solvers
are all evaluated under the same physical delays, inventory conservation, and
capacity limits.

\subsection{Reward and KPI Accounting}

The objective of the agent is to maximize network-wide profitability.
At time step~$t$, the reward~$r_t$ is the sum of realized period
profits over managed nodes. Let $M \subseteq V$ denote the set of
managed nodes (retailers, distributors, and factories, excluding
raw-material sources and end markets). Conditional on the realized
demand and feasible filled orders in period~$t$, the reward follows a
standard additive inventory cost decomposition~\citep{porteus2002}:
\begin{equation}\label{eq:reward}
\scriptsize
\begin{aligned}
r_t
&=\sum_{j\in M}P_t^{(j)}
=\alpha^t\sum_{j\in M}\Bigl[
\underbrace{\sum_{k\in\operatorname{succ}(j)}p_{jk}S_t^{(j,k)}}_{\mathrm{SR}}
-\underbrace{\sum_{i:(i,j)\in E_r}p_{ij}R_t^{(i,j)}}_{\mathrm{PC}}
-\underbrace{h_jX_{t+1}^{(j)}}_{\mathrm{HC}} \\
&\qquad
-\underbrace{\sum_{i:(i,j)\in E_r}g_{ij}Y_{t+1}^{(i,j)}}_{\mathrm{PHC}}
-\underbrace{\frac{o_j}{v_j}\sum_{k\in\operatorname{succ}(j)}S_t^{(j,k)}}_{\mathrm{OC}}
-\underbrace{\sum_{k:(j,k)\in E_d}b_{jk}U_t^{(j,k)}}_{\mathrm{SP}}
-\underbrace{\sum_{i:(i,j)\in E_r}K_{ij}\mathbf{1}\{R_t^{(i,j)}>0\}}_{\mathrm{FK}}
\Bigr].
\end{aligned}
\end{equation}
The bracket labels denote sales revenue~(SR), procurement cost~(PC),
node holding cost~(HC), pipeline holding cost~(PHC), operating cost~(OC),
shortage penalty~(SP), and fixed ordering cost~(FK). Here $p_{jk}$ is the
unit transfer or retail price on edge~$(j,k)$, $h_j$ is the node holding cost, $g_{ij}$ is
the inbound pipeline holding cost, $o_j$ is the operating-cost coefficient,
$v_j$ is the production yield, $b_{jk}$ is the per-unit shortage penalty,
and $K_{ij}$ is an optional fixed fee for a physically filled order on
edge~$(i,j)$. The factor $\alpha \in (0,1]$ is the environment's financial
profit-discount factor, distinct from any algorithm-specific discount used
internally during RL training. Fixed ordering costs default to zero unless
specified by the topology.

Because the reward is computed after demand realization and retail
fulfillment, holding and pipeline costs accrue on post-step stocks
$X_{t+1}$ and $Y_{t+1}$, while shortage penalties accrue on
end-of-period unfulfilled demand~$U_t$. The fixed ordering cost is
triggered by $\mathbf{1}\{R_t^{(i,j)} > 0\}$, not by the raw request
$a_t^{(i,j)}$; therefore a controller is not charged a setup fee for
a request that cannot be physically filled because of upstream inventory,
yield, or capacity constraints. The same decomposed terms are exported as
benchmark KPIs, together with service level, fill rate, unfulfilled
demand, average inventory, and bullwhip diagnostics.

\subsection{Demand Processes and Goodwill Feedback}

The environment extends classical stationary Poisson demand by
supporting composable non-stationary perturbations: external empirical
demand traces, linear trends, sinusoidal seasonality, step-function
shocks, and noise scaling around the current mean. The demand engine can
also accept raw time-series vectors, such as historical retail demand
profiles, as the base realization path. Benchmark scenarios can therefore
stress both predictable parametric non-stationarity (trend and seasonality)
and highly irregular empirical regime shifts.

Additionally, we introduce an endogenous customer goodwill dynamic,
motivated by service-dependent demand and goodwill-loss models in inventory
theory~\citep{schwartz1966,olsen2008}. When activated, the service level
establishes an asymmetric feedback loop:
end-of-period unfulfilled demand~$U_t$ degrades customer sentiment~$s_t$,
thereby scaling the mean of the next demand draw. The sentiment state
evolves as:
\begin{equation}\label{eq:goodwill}
  s_{t+1} =
  \begin{cases}
    \max\bigl(s_{\min},\;\gamma_{\text{drop}} \cdot s_t\bigr)
      & \text{if stockout at } t,\\
    \min\bigl(s_{\max},\;\gamma_{\text{rec}} \cdot s_t\bigr)
      & \text{otherwise,}
  \end{cases}
\end{equation}
where $\gamma_{\text{drop}} = 0.90$, $\gamma_{\text{rec}} = 1.01$,
$s_{\min} = 0.2$, and $s_{\max} = 2.0$. Here, a stockout event is
triggered when aggregate unfulfilled demand across retail-demand links is
strictly positive, $\sum_{(j,k)\in E_d} U_t^{(j,k)} > 0$. The effective
demand mean for the next draw is
$\lambda_{t+1}=s_{t+1}\bar{\lambda}_{t+1}$, where
$\bar{\lambda}_{t+1}$ is the exogenous mean after any trend, seasonal,
shock, or external-series effects. Because sentiment decays abruptly
but recovers slowly (\Cref{fig:goodwill_dynamics}), goodwill creates
path-dependent non-stationarity: the same exogenous demand profile can
generate different future demand trajectories depending on the policy's
past service failures. This provides a stress test for both learned
policies and rolling-horizon OR baselines, because endogenous demand
feedback couples present fulfillment decisions to future market size.

\begin{proposition}[Goodwill Drift Threshold]\label{prop:goodwill}
  Consider a stationary service process with long-run no-stockout
  probability~$\rho = \Pr(\text{no stockout at }t)$. Ignoring the
  reflecting floor and cap, the expected log-drift of goodwill is
  non-negative when
  \[
    \rho \;\geq\; \rho_{\min}
      \;=\; \frac{\ln(1/\gamma_{\text{drop}})}
                  {\ln(1/\gamma_{\text{drop}})
                   + \ln(\gamma_{\text{rec}})}
      \;\approx\; 91.37\%.
  \]
  Below this threshold, goodwill has negative multiplicative drift and
  is pushed toward the floor~$s_{\min}$; above it, goodwill tends to
  recover toward the cap~$s_{\max}$.
\end{proposition}

This threshold is not an additional assumption in the simulator; it is
a diagnostic interpretation of the asymmetric update rule under a
stationary service process. It formalizes why goodwill scenarios sharply
punish repeated service failures: repeated short-run stockouts can
create persistent demand erosion unless future service reliability is
high enough to offset the decay.

\section{Experimental Setup and Baseline Analysis}\label{sec:results}

This section instantiates the benchmark contract defined in
\Cref{sec:framework}. We evaluate the released scenario matrix across
topology, demand regime, shortage treatment, goodwill feedback,
information access, and controller family. The evaluation is organized
around a canonical 22-scenario core grid, with four supplemental
MARL-mode scenarios reported separately in the same merged artifact,
yielding a 26-scenario audit matrix over 10 canonical seeds. The
benchmark includes 29 registered non-LLM agent configurations (Oracle,
rolling-horizon OR, heuristics, RL, hybrid, and imitation-learning
variants) plus the full-matrix LLM-Policy-C baseline. Neural policies are
implemented in PyTorch~\citep{paszke2019} and trained using
Stable-Baselines3~\citep{raffin2021}; all methods are evaluated through
the same \texttt{CoreEnv} transition, reward, action-bound, and KPI
contract.

\subsection{Evaluated Solution Families}

\Cref{tab:paradigm_comparison} maps the evaluated roster into solution
families and demand-visibility tiers. Following the information protocol
defined in \Cref{sec:information_protocol}, architectural inductive bias
is reported separately from access to demand-context information. This
distinction is important because demand visibility can change both the
available decision information and the optimization landscape faced during
training; separating it from architecture prevents an informed model from
being mistaken for an intrinsically better architecture. For learned
policies, informed and blind variants are trained as separate checkpoints;
information access is therefore treated as an experimental factor rather
than assumed to provide monotonic improvement.

The roster consists of 29 registered non-LLM agent IDs plus the
full-matrix LLM-Policy-C baseline. The Oracle is reported as a non-causal
reference: a perfect-information LP in exogenous-demand settings and a
clairvoyant simulation-optimization benchmark under endogenous goodwill.
Rolling-horizon OR baselines are represented by MSSP and DLP; classical
heuristics provide interpretable industrial references; learned families
cover flat-vector PPO/SAC, graph-message-passing PPO, Transformer PPO,
residual RL, and imitation-learning variants. The LLM-Policy-C baseline
is included as a bounded policy-parameter diagnostic for foundation-model
reasoning, not as a replacement for specialized inventory controllers.
Diagnostic PPO-MLP-v1 and PPO-MLP-raw ablations are retained in the
released CSV but omitted from the condensed main tables.
For reproducibility, the released CSV also preserves a few legacy
implementation prefixes: \texttt{GNN\_V3} and \texttt{GNN\_V3\_B}
correspond to PPO-GNN and PPO-GNN-B, respectively.

\begin{table}[!htbp]
\centering
\caption{Paradigm comparison of evaluated solution families. The
canonical benchmark registers 29 non-LLM agent IDs (including Oracle
and ablations) plus the full-matrix LLM-Policy-C baseline. Direct
per-period LLM prompting variants are retained as diagnostics in the
appendix.}
\label{tab:paradigm_comparison}
\footnotesize
\renewcommand{\arraystretch}{1.05}
\setlength{\tabcolsep}{4pt}
\begin{tabular}{@{} l l l p{5.5cm} @{}}
\toprule
\textbf{Agent}
  & \textbf{Paradigm}
  & \makecell[l]{\textbf{Information}\\\textbf{Access}}
  & \makecell[l]{\textbf{Solution}\\\textbf{Protocol}} \\
\midrule
\rowcolor{lightgray}
Oracle      & Bound       & Non-causal       & Clairvoyant LP / fixed-point benchmark \\
MSSP / MSSP-I
            & SAA-RH OR   & Blind / Informed & Multi-stage stochastic program \\
\rowcolor{lightgray}
DLP / DLP-I & RH OR      & Blind / Informed  & Deterministic expected-value LP \\
Newsvendor / NV-I
            & Heuristics    & Blind / Informed  & Critical-Ratio~\citep{arrow1951} \\
\rowcolor{lightgray}
$(s,S)$ / $(s,S)$-I
            & Heuristics    & Blind / Informed  & Reorder-point/order-up-to~\citep{scarf1960ss} \\
ExpSmooth / ExpSmooth-I
            & Adapt.\ Heur. & Blind / Informed  & Holt linear smoothing \\
\rowcolor{lightgray}
Echelon / Echelon-I
            & Heuristics    & Blind / Informed  & Echelon-inspired base-stock~\citep{clark1960} \\
\midrule
\rowcolor{lightgray}
DAgger-B / DAgger-G
            & Imitation L.  & Blind / Informed & Dataset Aggregation~\citep{ross2011} \\
GNN-IL      & Imitation L.  & Graph states      & Behavioral cloning + PPO fine-tuning \\
\rowcolor{lightgray}
PPO-GNN / PPO-GNN-B
            & Advanced RL   & Informed / Blind  & Directed MPNN + PPO~\citep{schulman2017,gilmer2017,velickovic2018} \\
\rowcolor{lightgray}
PPO-Transformer
            & Advanced RL   & Informed          & Node-Token Transformer + PPO~\citep{vaswani2017,schulman2017} \\
ST-PPO / ST-PPO-B
            & Advanced RL   & Informed / Blind  & Spatio-Temporal Transformer + PPO~\citep{vaswani2017,schulman2017,zambaldi2018,parisotto2020} \\
SAC / SAC-B & Deep RL       & Informed / Blind  & Soft Actor-Critic~\citep{haarnoja2018} \\
\rowcolor{lightgray}
Residual / Res.-B
            & Hybrid        & Informed / Blind  & PPO + Heuristic~\citep{silver2019} \\
PPO-MLP / PPO-MLP-B
            & Standard RL   & Informed / Blind  & PPO~\citep{schulman2017} \\
\rowcolor{lightgray}
LLM-Policy-C
            & Foundation    & Episode-level prompt & LLM strategy + deterministic controller~\citep{quan2024} \\
\bottomrule
\end{tabular}
\end{table}

\textbf{Oracle (upper bound / clairvoyant heuristic).}
For exogenous demand, the Oracle is a perfect-information
linear/mixed-integer program that observes the realized demand sequence
before selecting a plan. The model is written in PuLP, a Python
linear-programming modeling package, and solved with COIN-OR
Branch-and-Cut (CBC)~\citep{mitchell2011,forrest2005cbc} against the
same inventory, pipeline, backlog, lead-time, capacity, and reward
contract as the online agents. For endogenous goodwill, future demand is
policy-dependent, so the Oracle is not a certified dynamic-programming
optimum. We instead report it as a clairvoyant simulation-optimization
benchmark using a bidirectional fixed-point procedure over pessimistic
and optimistic demand traces.

\textbf{Rolling-horizon LP baselines (MSSP, DLP).}
The Multi-Stage Stochastic Program~(MSSP)~\citep{birge2011} is
implemented as a rolling-horizon sample-average approximation with
non-anticipativity constraints. MSSP-I uses the informed future demand
mean path, while MSSP estimates future demand from the causal history.
The Deterministic LP~(DLP) replaces the scenario tree with an
expected-value trajectory, producing a cheaper but risk-neutral rolling
baseline. Both solvers cap their horizon at the remaining episode
length and re-solve online from the current simulator state.

\textbf{Heuristics.}
The classical baselines are centralized, graph-aware adaptations rather
than exact implementations of single-item textbook policies. Newsvendor
uses a critical-ratio base-stock target~\citep{arrow1951,porteus2002};
$(s,S)$ uses a reorder-point/order-up-to batching rule~\citep{scarf1960ss,porteus2002};
ExpSmooth uses Holt linear smoothing~\citep{hyndman2021}; and
EchelonApprox is a Clark--Scarf-inspired echelon base-stock
approximation~\citep{clark1960}, not an exact dynamic program. Informed
variants use current demand parameters or variance information where
available; blind variants estimate from realized history.

\textbf{Imitation learning.}
DAgger-B and DAgger-G~\citep{ross2011} evaluate GNN-based policies
trained from expert demonstrations under blind and informed observation
contracts, respectively. GNN-IL provides a separate
behavioral-cloning-plus-PPO-finetuning diagnostic using the same
directed graph feature extractor as PPO-GNN. We treat these as
imitation-learning baselines and diagnostics for distribution shift,
not as direct claims that imitation dominates the RL or OR families.

\textbf{Deep RL and hybrid agents.}
All RL agents are trained using Stable-Baselines3~\citep{raffin2021}.
PPO-MLP uses standard Proximal Policy Optimization~\citep{schulman2017}
with a flat MLP policy.
\textbf{PPO-GNN} replaces the MLP feature extractor with a directed
edge-conditioned Message Passing Neural Network
(MPNN)~\citep{gilmer2017,velickovic2018} that processes
the supply chain graph via bidirectional (upstream/downstream)
message passing with edge features, attention-weighted aggregation,
residual connections, LayerNorm, and BatchNorm.
We introduce two Transformer-based PPO variants:
(i)~\textbf{PPO-Transformer}, which treats each supply-chain node as a token
and applies spatial self-attention~\citep{vaswani2017}; and
(ii)~\textbf{ST-PPO}, which extends PPO-Transformer with temporal frame
stacking ($n{=}4$), creating $(\text{node} \times \text{timestep})$ tokens
for joint spatio-temporal attention~\citep{parisotto2020}.
SAC uses Soft Actor-Critic~\citep{haarnoja2018}. Residual RL adds a learned
correction to a base-stock heuristic~\citep{silver2019}.
Architecture details and hyperparameters are reported in the full appendix
version.

\textbf{Foundation-model baseline.}
To test whether a local foundation model can contribute useful
inventory decisions without being queried at every period, LLM-Policy-C
uses a Qwen2.5-1.5B-Instruct model once per episode to select bounded
policy parameters for a deterministic controller. This ``LLM as
strategist, code as controller'' setup is the only LLM variant included
in the full 26-scenario matrix. Direct raw-action prompting and
InvAgent-inspired staged prompts are retained as diagnostics in the full
appendix version because they suffer from high latency and action-format
failures.

\textbf{Generalist training protocol.}
For the main PPO, SAC, Transformer, GNN, and Residual RL controllers,
each reported checkpoint is a \emph{single generalist model per
(architecture~$\times$~topology)}, not a per-scenario specialist.
During RL training, the topology is fixed (base or serial), while
demand profile, goodwill, fulfillment mode, and noise scale are
randomized at every episode reset via domain randomization. This
produces one checkpoint per topology (e.g.,
\texttt{ppo-transformer\_base.zip}, \texttt{st-ppo\_serial.zip}).
The imitation-learning diagnostics are also evaluated from fixed
topology-level checkpoints: DAgger-G/B use graph imitation policies,
while GNN-IL uses Oracle behavioral cloning followed by PPO fine-tuning
on the canonical stationary training environment. Thus, GNN-IL should be
read as an imitation diagnostic rather than as part of the
domain-randomized RL training pool.

At evaluation time, each learned checkpoint is tested on the full
22-scenario fixed grid across 10~canonical seeds (220~episodes per
agent), using deterministic inference. Four additional MARL-mode
scenarios are evaluated separately (documented in the full appendix
version), bringing the total to 26~scenarios (260~episodes). This design
explicitly imposes a \emph{generalization tax}: learned policies must
deploy across diverse regimes without scenario-specific retraining,
while OR solvers and heuristics adapt online to the current state and,
in informed variants, to current demand parameters. Reported neural
performance therefore reflects cross-regime deployment robustness, not
scenario-tuned optimality.

\subsection{Scenario Matrix and Evaluation Protocol}

The main benchmark uses a curated 22-scenario core matrix
(\Cref{tab:scenario_axes}). The design is intentionally representative
rather than exhaustive: it combines topology, demand regime, endogenous
feedback, and fulfillment mode to cover stress patterns common in the
inventory-control literature while remaining small enough for
cross-paradigm evaluation over many agents and seeds. Starting from the
conceptual factorial design
($2$~topologies $\times$ $4$~demand regimes $\times$ $2$~goodwill modes
$\times$ $2$~fulfillment modes $=32$), we apply two exclusions. First,
stationary and M5 external-trace profiles are not paired with endogenous
goodwill, because goodwill is designed to stress service-dependent
non-stationarity ($-8$). Second, the M5 trace is evaluated only under
backlog fulfillment, avoiding an interpretation of observed retail
sales as an uncensored lost-sales demand process ($-2$).
The M5 scenario is derived from the real \texttt{sales\_train\_evaluation}
item-store panel: we select a 30-day item-store window with high rolling
coefficient of variation subject to minimum activity filters, scale the
observed unit-sales path to the benchmark mean demand level
($\bar d=20$), and replay it deterministically as an external demand
trace. This makes the scenario a real-data stress test rather than a
claim that one M5 item is representative of all Walmart demand. The
canonical M5 rows use the M5 panel for terminal demand only: calendar
features, SNAP/event indicators, sell-price histories, and Walmart's
unobserved replenishment network are not treated as observed supply-chain
physics. Lead times, capacities, inventory costs, and shortage penalties
remain controlled benchmark parameters. A separate adapter can infer a
hierarchy-based planning DAG from M5 metadata, but such hierarchy-inferred
topologies change the action space and are reported as custom topology
experiments rather than replacements for the canonical base/serial M5 rows.

The resulting 22 core scenarios consist of 16 synthetic non-stationary
stress tests, four stationary paper-replication scenarios, and two
M5-derived external-trace scenarios. Four additional MARL-mode rows
reuse non-goodwill backlog synthetic scenarios with the MARL evaluation
flag enabled; these supplemental rows bring the merged artifact to
26~scenarios and are reported separately where relevant. The core and
supplemental rows are enumerated in the full appendix version.

\begin{table}[!htbp]
\centering
\caption{Core scenario axes for the 22-scenario evaluation matrix. The
four supplemental MARL-mode rows use the same topology and demand axes
but are not part of the core 22-scenario aggregate unless explicitly
stated.}
\label{tab:scenario_axes}
\footnotesize
\renewcommand{\arraystretch}{1.05}
\begin{tabular}{@{} >{\bfseries}l l p{5.2cm} @{}}
\toprule
Axis & Configurations & Benchmarking Purpose \\
\midrule
\rowcolor{lightgray}
Topology\,(2)
  & \makecell[tl]{Default network (9-node),\\Serial (5-node)}
  & Contrasts branching/merging allocation and capacity interactions
    with linear pipeline delays. \\
Demand\,(4)
  & \makecell[tl]{Stationary, M5-derived volatile,\\Trend+Seasonal,\\Trend+Seas.+Shock}
  & Tests stationary control, external-trace robustness, and composable
    non-stationary distribution shifts. \\
\rowcolor{lightgray}
Dynamics\,(2)
  & \makecell[tl]{Exogenous,\\Endogenous Goodwill}
  & Evaluates resilience against service-level feedback loops in
    synthetic non-stationary settings. \\
Fulfillment\,(2)
  & \makecell[tl]{Backlog,\\Lost Sales}
  & Shifts the penalty landscape from accumulated debt to immediate loss. \\
\bottomrule
\end{tabular}
\end{table}

\subsection{Main Benchmark Results and Compute Trade-off}

\Cref{tab:summary_performance} consolidates the 22-scenario core
benchmark into three interpretable regimes: stationary demand,
non-stationary exogenous demand, and endogenous-goodwill demand. The
dominant quality result is that informed stochastic programming
(MSSP-I) is the strongest causal non-Oracle baseline, reaching
approximately \textbf{95\%} of Oracle profit on the core grid. This is not a claim
that stochastic programming is universally deployment-friendly: MSSP-I
requires repeated rolling-horizon solves and informed forecast access.
Rather, it provides the empirical quality ceiling among causal methods
in this benchmark.

The learned-policy frontier is different. PPO-Transformer is the
strongest fast learned policy in aggregate, reaching approximately
\textbf{75\%} of Oracle across the core grid, while Residual~RL remains
close at approximately \textbf{73\%}. DAgger-B is highly effective
in-distribution, capturing approximately \textbf{95\%} of Oracle
under stationary demand, but its non-stationary performance falls to
33\%, illustrating imitation learning's sensitivity to distribution
shift. SAC is a cautionary counterexample: the informed checkpoint has
negative stationary performance on average and is dominated by its blind
counterpart, indicating checkpoint instability rather than a monotone
benefit from richer observations. GNN-IL is therefore retained as a
diagnostic imitation baseline rather than a headline method. Finally, the full-matrix LLM-Policy-C
baseline reaches approximately \textbf{60\%} of Oracle on the core grid:
competitive
with several classical and neural baselines, but with much higher
latency than specialized learned policies.

\Cref{tab:compute_quality} reports the corresponding computational
costs. These timings should be interpreted as evaluation latency on the
reported hardware, not universal wall-clock constants; they mainly show
the amortization pattern whereby trained policies replace repeated
optimization with a forward pass or a lightweight heuristic correction.

\begin{table*}[!htbp]
\caption{Core performance and computational cost by paradigm. Performance
  columns report scenario-wise \% of Oracle profit, averaged over the
  22-scenario core grid (mean $\pm$ sample cross-scenario SD; these are
  heterogeneity summaries, not seed-level confidence intervals). Inference
  time is mean seconds per episode on the same hardware; speed-up is
  relative to blind MSSP.}
\label{tab:perf_and_compute}
\label{tab:summary_performance}
\label{tab:compute_quality}
\centering
\scriptsize
\renewcommand{\arraystretch}{1.05}
\setlength{\tabcolsep}{2.6pt}
\begin{tabular}{@{} l l rrrr rr @{}}
\toprule
\textbf{Category} & \textbf{Agent}
  & \makecell{\textbf{Stat.}\\\textbf{(4)}}
  & \makecell{\textbf{Non-S.}\\\textbf{(10)}}
  & \makecell{\textbf{End.}\\\textbf{(8)}}
  & \textbf{All}
  & \makecell{\textbf{s/}\\\textbf{ep}}
  & \textbf{Speed} \\
\midrule
\rowcolor{lightgray}
\textit{Bound} & Oracle & 100 & 100 & 100 & \textbf{100} & 0.84 & --- \\
\midrule
\multirow{4}{*}{\textit{Rolling-Horizon OR}}
  & MSSP             &  92$\pm$5  &  61$\pm$26 &  58$\pm$7  &  66$\pm$22 & 10.28 & 1$\times$ \\
  & MSSP-I           &  97$\pm$1  &  99$\pm$1  &  89$\pm$3  &  \textbf{95}$\pm$5 & 7.68 & 1.3$\times$ \\
  & DLP              &  78$\pm$15 &  36$\pm$29 &  52$\pm$11 &  49$\pm$26 & 1.06 & 9.7$\times$ \\
  & DLP-I            &  80$\pm$15 &  50$\pm$32 &  58$\pm$18 &  58$\pm$27 & 1.00 & 10.2$\times$ \\
\midrule
\multirow{5}{*}{\textit{Heuristics}}
  & Newsvendor       &  83$\pm$7  &  46$\pm$32 &  56$\pm$7  &  56$\pm$25 & 0.049 & 210$\times$ \\
  & $(s,S)$          &  86$\pm$2  &  51$\pm$32 &  61$\pm$7  &  61$\pm$25 & 0.032 & 317$\times$ \\
  & ExpSmooth        &  84$\pm$5  &  45$\pm$32 &  55$\pm$7  &  56$\pm$26 & 0.037 & 281$\times$ \\
  & Echelon          &  84$\pm$5  &  50$\pm$43 &  55$\pm$6  &  58$\pm$31 & 0.058 & 176$\times$ \\
  & Echelon-I        &  88$\pm$3  &  67$\pm$17 &  71$\pm$6  &  72$\pm$14 & 0.054 & 192$\times$ \\
\midrule
\multirow{2}{*}{\textit{Imit.\ L.}}
  & DAgger-B         &  95$\pm$2  &  33$\pm$25 &  59$\pm$12 &  53$\pm$29 & 0.273 & 38$\times$ \\
  & GNN-IL           &  84$\pm$7  &  3$\pm$54  &  25$\pm$26 &  26$\pm$48 & 0.275 & 37$\times$ \\
\midrule
\multirow{5}{*}{\textit{Deep RL}}
  & PPO-MLP          &  77$\pm$9  &  56$\pm$35 &  62$\pm$23 &  62$\pm$28 & 0.132 & 78$\times$ \\
  & PPO-GNN          &  73$\pm$3  &  57$\pm$35 &  62$\pm$23 &  61$\pm$27 & 0.305 & 34$\times$ \\
  & PPO-Transformer  &  74$\pm$6  &  77$\pm$13 &  71$\pm$12 &  75$\pm$12 & 0.229 & 45$\times$ \\
  & ST-PPO           &  69$\pm$2  &  59$\pm$32 &  64$\pm$20 &  63$\pm$24 & 0.269 & 38$\times$ \\
  & SAC              &  -13$\pm$53 &  35$\pm$47 &  73$\pm$13 &  40$\pm$49 & 0.135 & 76$\times$ \\
\midrule
\textit{Hybrid} & Residual &  87$\pm$4  &  68$\pm$17 &  73$\pm$6  &  73$\pm$14 & 0.223 & 46$\times$ \\
\midrule
\textit{Foundation} & LLM-Policy-C &  77$\pm$2  &  54$\pm$22 &  58$\pm$17 &  60$\pm$20 & 8.150 & 1.3$\times$ \\
\bottomrule
\end{tabular}
\end{table*}

\paragraph{Compute--quality interpretation.}

\Cref{tab:compute_quality} shows that solution quality and latency do
not move together monotonically. MSSP-I provides the strongest causal
quality benchmark, but blind MSSP still requires roughly
10.3~seconds per 30-period episode on the reported hardware because it
re-solves a sampled scenario tree online. DLP is faster
(${\sim}$1.1~s/episode) but sacrifices scenario hedging and performs
substantially worse under stochastic non-stationarity.

Trained neural policies amortize this repeated optimization into
offline training plus fast evaluation. PPO-Transformer reaches
approximately 75\% of Oracle profit while running in
${\sim}$0.229~s/episode, about 45$\times$ faster than blind MSSP.
PPO-MLP is also lightweight (${\sim}$0.132~s/episode) but gives up some
quality, while graph and residual policies incur additional feature
extraction or heuristic-evaluation overhead. Residual~RL is the
strongest hybrid policy in the table by scenario-wise percentage
(73\% of Oracle), while its inference latency
(${\sim}$0.223~s/episode) is comparable to the Transformer policies and
well below DLP.

Training time should therefore be interpreted as an offline amortized
cost, not as part of every episode's latency. For a learned policy used
over $N$ future episodes, the effective computational cost is
$C_{\mathrm{train}}/N + C_{\mathrm{infer}}$. Training costs are not
included in the compact table because several legacy pretrained
artifacts lack retained training logs; those artifacts are evaluated from
shipped weights but are not used for exact training-time claims. This
favors OR solvers and heuristics for one-off planning, but favors
reusable generalist policies when the same trained controller is deployed
repeatedly across stores, SKUs, seeds, or rolling operational cycles.
Specialist policies face a higher effective cost whenever a separate
checkpoint must be trained for each scenario.

The LLM-Policy-C baseline occupies a different point on this frontier:
it avoids per-period language-model calls by querying once per episode,
but still averages roughly 8.2~s/episode. This is slightly faster than
blind MSSP in the final artifact but far slower than DLP, heuristics, and
specialized learned policies. Its role is therefore diagnostic and
strategic rather than latency-competitive. Overall, the
compute results support an amortization narrative, not a universal
neural dominance claim: specialized learned policies can be much faster
than rolling-horizon stochastic programming, but the magnitude depends
strongly on architecture and whether the controller invokes graph,
heuristic, or language-model components.

\subsection{Information Access and Architecture Effects}

Where paired variants are available, we evaluate agents under a common
\textbf{information parity protocol}. Blind variants estimate demand
from observation history, while informed variants receive the current
scenario's demand parameters through the same environment-native feature
contract. For learned policies, blind and informed results should be
read as separate trained checkpoints rather than as a guaranteed
monotone improvement from adding features: the additional information
can also change optimization dynamics and generalization.

\Cref{tab:info_value} shows that the value of information is largest
when the downstream algorithm can use it structurally. MSSP gains
29.2~percentage points because its scenario tree is centered on the
correct demand process; DLP gains only 9.2~points because its
expected-value LP still ignores variance. Classical heuristics benefit
when their target calculations can use demand-parameter information
directly, with double-digit gains of 14.1~points for Newsvendor,
14.0~points for EchelonApprox, and 10.0~points for $(s,S)$. In contrast,
the reactive exponential smoothing baseline gains only 0.7~points.

The learned-policy pairs are less monotonic. Residual RL improves by
15.2~points when informed features are available, while PPO-GNN and
ST-PPO change by roughly one point on the core grid. SAC is the
counterexample: its blind checkpoint exceeds the informed checkpoint by
19.9~points, driven by poor serial-topology training outcomes for the
informed SAC model. Thus, information access is an essential control
dimension, but the empirical ranking reflects an interaction among
information, architecture, and training stability rather than
information alone.

\begin{table}[!htbp]
\centering
\caption{Value of information for selected paired methods
  (scenario-wise \% of Oracle profit, averaged over the 22 core scenarios).}
\label{tab:info_value}
\small
\renewcommand{\arraystretch}{1.05}
\begin{tabular}{@{} l r r r @{}}
\toprule
\textbf{Agent}
  & \textbf{Blind (\%)}
  & \textbf{Informed (\%)}
  & \textbf{$\Delta$ (pp)} \\
\midrule
MSSP          & 65.7 & 94.9 & $+$29.2 \\
DLP           & 49.2 & 58.4 & $+$9.2 \\
Newsvendor    & 56.2 & 70.3 & $+$14.1 \\
$(s,S)$       & 61.1 & 71.1 & $+$10.0 \\
ExpSmooth     & 55.8 & 56.5 & $+$0.7 \\
Echelon       & 58.4 & 72.4 & $+$14.0 \\
Residual      & 58.0 & 73.2 & $+$15.2 \\
SAC           & 59.7 & 39.9 & $-$19.9 \\
\bottomrule
\end{tabular}
\end{table}

\paragraph{Topology-transfer summary.}
We treat topology transfer as an architectural stress test rather than
part of the main 22-scenario ranking. A Residual GCN-Pool policy trained
on the default 9-node topology retains positive profit when deployed
zero-shot on four unseen graphs, preserving 41--84\% of the Newsvendor
heuristic's profit depending on structural similarity. However, the
transfer experiments also expose a service-level failure: the
fixed-size action head can generate profitable inventory behavior while
failing to route material reliably to new retail edges. For this reason,
the full transfer ladder, statistical tests, and operational KPI
decomposition are reported in the full appendix version; the main text
uses the result primarily to motivate topology-aware action decoding.

\subsection{Stress Tests and Foundation-Model Policy Baseline}

The stress-test regimes explain why the aggregate ranking should not be
read as a single universal ordering. Under non-stationary exogenous
demand, MSSP-I remains near the Oracle because its scenario tree hedges
against future demand paths, while DLP-I can under-stock sharply on
serial topologies because it optimizes only an expected-value path.
DAgger-B provides the opposite lesson: it reaches 94.6\% of Oracle in
stationary regimes but drops under trend, seasonality, and shocks,
consistent with imitation learning's sensitivity to covariate shift.

Endogenous goodwill creates a different stressor because future demand
depends on the service path induced by the policy. In the eight goodwill
core scenarios, blind MSSP averages 57.9\% of Oracle, while MSSP-I
recovers to 89.1\% by centering its scenario tree on the informed demand
path. The strongest learned and hybrid policies remain meaningful but
below MSSP-I: Residual~RL reaches 73.0\%, SAC 72.8\%, and
PPO-Transformer 71.4\% of Oracle. These results indicate that learned
policies can buffer against service-feedback dynamics, but they do not
replace informed stochastic programming in this benchmark.
They are also consistent with the drift interpretation in
\Cref{prop:goodwill}: goodwill rows remain severe because even policies
with high fill rates can experience no-stockout probabilities below the
91.37\% recovery threshold, so occasional service failures continue to
exert downward pressure on sentiment.

The foundation-model experiments add a final stress test around action
formatting and latency. Direct raw-action prompting with the local
\texttt{Qwen2.5-1.5B-Instruct} model~\citep{qwen2024} produced over-ordering,
parse/shape failures, and multi-minute episodes. The full-matrix
LLM-Policy-C variant instead queries the model once per episode for
bounded base-stock multipliers and delegates execution to deterministic
controllers. This stabilizes the action format and reaches about 60\%
of Oracle on the core grid. It averages roughly 8.2~seconds per episode
on the core grid and about 9.0~seconds over all 26 rows, making it
slightly faster than blind MSSP but far slower than DLP, heuristics, and
specialized learned policies. LLMs are
therefore best interpreted here as strategic policy-parameter generators
rather than high-frequency inventory controllers.

\begin{figure}[!htbp]
  \begin{subfigure}[t]{0.48\linewidth}
    \centering
    \includegraphics[width=\linewidth]{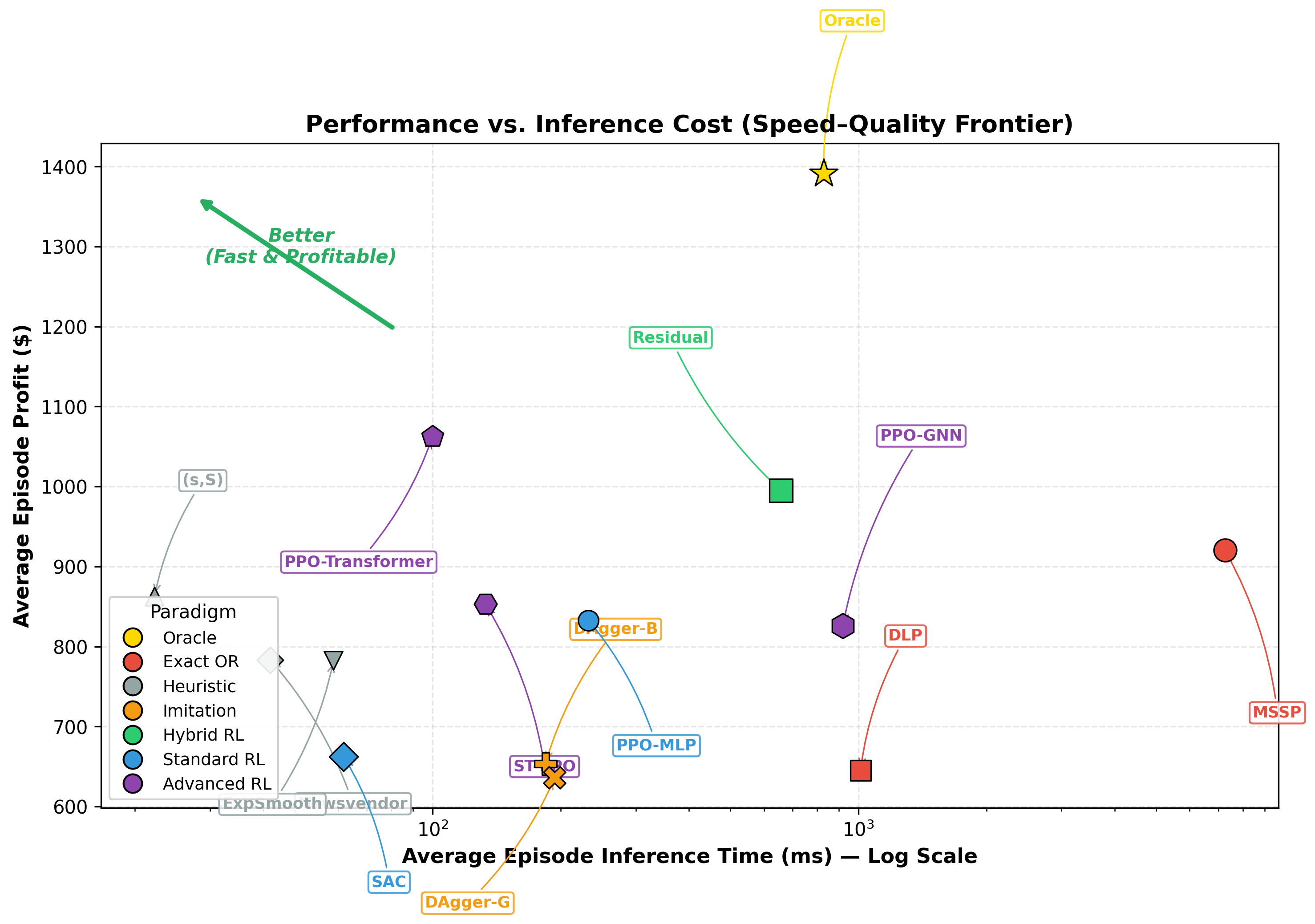}
    \caption{Speed--quality Pareto frontier.}
    \label{fig:speed_quality}
  \end{subfigure}
  \hfill
  \begin{subfigure}[t]{0.48\linewidth}
    \centering
    \includegraphics[width=\linewidth]{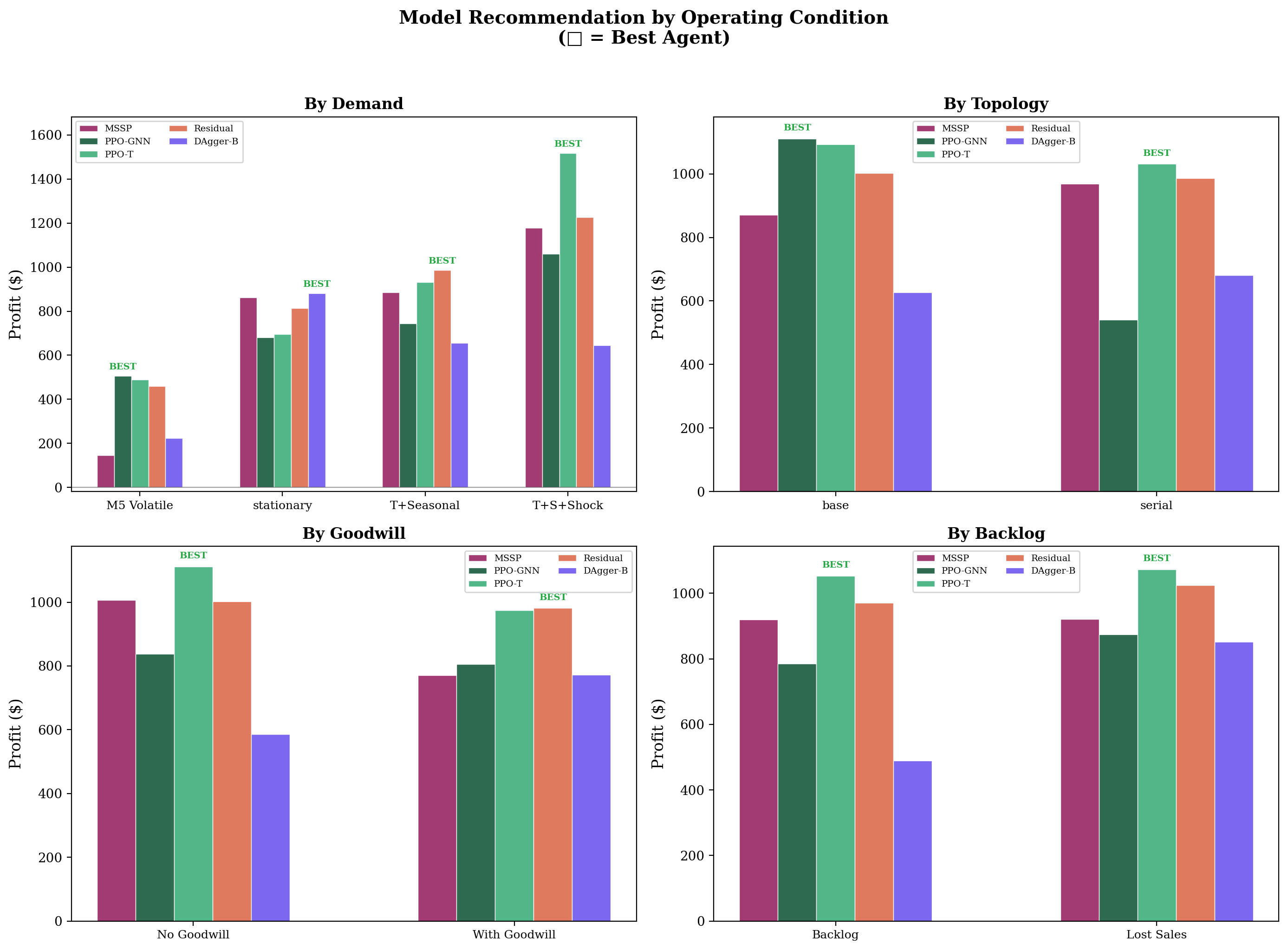}
    \caption{Model recommendation matrix.}
    \label{fig:recommendation_matrix}
  \end{subfigure}
  \caption[Operational characteristics: speed vs.\ quality and scenario-level recommendations.]{%
    \emph{(a)}~Average episode profit vs.\ inference time (log scale), illustrating the separation between rolling-horizon OR quality, fast learned-policy inference, and the slower LLM-policy baseline.
    \emph{(b)}~Model recommendation by operating condition. Four panels decompose mean profit by demand regime, topology, goodwill mode, and fulfillment policy. ``BEST'' labels mark the top non-Oracle agent.}
  \label{fig:operational_characteristics}
\end{figure}

\Cref{fig:speed_quality} visualizes the speed--quality
frontier, while \Cref{fig:recommendation_matrix} supports method
selection by decomposing performance along each scenario axis. The full
appendix version provides scenario-level heatmaps, training curves,
transfer diagnostics, and detailed LLM diagnostics.

\section{Discussion, Limitations, and Future Directions}\label{sec:conclusion}

\subsection{Discussion}

The central lesson from \texttt{gym-invmgmt} is not that one paradigm
dominates inventory control. Instead, the benchmark exposes a
deployment trade-off among information access, uncertainty modeling,
architectural bias, and computational amortization. Informed stochastic
programming provides the strongest causal non-Oracle quality benchmark:
MSSP-I achieves 94.9\% of Oracle on the 22-scenario core grid. Its advantage over DLP-I shows that
the value of OR in this setting is not merely optimization over a
network flow model, but explicit scenario hedging under stochastic
non-stationarity.

Learned policies occupy a different part of the frontier. PPO-Transformer
and Residual~RL reach 73--75\% of Oracle on the core
grid, below MSSP-I but with substantially lower online latency. This
does not support a blanket claim that neural policies replace
rolling-horizon solvers. It supports a more specific claim: once trained,
generalist neural controllers can amortize decision making across many
episodes, stores, seeds, or rolling deployments, making them attractive
when repeated online optimization is too slow or operationally
cumbersome. The strongest learned methods are therefore best viewed as
fast reusable approximations, not as universal substitutes for informed
stochastic programming.

The benchmark also shows why information access must be treated as an
experimental factor rather than a slogan. Informed demand parameters
help MSSP substantially and provide double-digit gains for
variance-sensitive heuristics such as Newsvendor, EchelonApprox, and
$(s,S)$, but they do not guarantee monotonic gains for learned agents.
SAC, for example, is worse in its informed variant because checkpoint
quality and optimization stability interact with the feature contract.
This is a useful negative result: inventory benchmarks should report
blind and informed variants explicitly, because additional information
can change both what the policy knows and how difficult the policy is to
train.

Finally, the diagnostic baselines clarify failure modes that aggregate
profit alone would hide. DAgger-B imitates Oracle-like behavior under
stationary and M5 external-trace demand, but degrades under
trend/seasonal shock regimes, illustrating covariate shift. GNN-IL is
therefore retained as a diagnostic BC-to-PPO pipeline rather than a
headline method. LLM-Policy-C is stable enough for the full scenario
matrix when used as a low-frequency policy-parameter generator, but it
remains slower and weaker than the best OR, heuristic, and learned
controllers; direct per-period LLM action generation is still brittle.
The topology-transfer experiments similarly show partial profit
transfer but service-level collapse, suggesting that graph encoders
alone are insufficient without topology-native action decoders.

\subsection{Methodological Limitations}

The results should be read as controlled benchmark evidence rather than
as universal rankings. The first limitation is \textbf{scenario and
topology scope}: the 22-scenario core grid covers patterns that are
common in the literature---serial and default multi-echelon topologies,
stationary and non-stationary demand, backlog and lost-sales fulfillment,
and endogenous goodwill feedback---but it is not a statistically
representative sample of all supply chains. The value of the grid is
therefore comparative control: it shows how method rankings change when
specific assumptions are varied under identical simulator mechanics.

A second limitation concerns \textbf{external data mapping}. The M5 rows
inject a real item-store sales trace as terminal demand, but they do not
reconstruct Walmart's physical replenishment network. Calendar features,
SNAP/event indicators, sell-price histories, upstream lead times,
capacities, and costs are not treated as observed logistics variables in
the canonical M5 rows. The released adapters can infer planning DAGs from
public retail hierarchies, but those graphs should be interpreted as
dataset-derived stress tests unless external logistics data are added.

The evaluation also separates \textbf{seed-level evaluation uncertainty}
from full training-run uncertainty. Each scenario is evaluated over 10
canonical seeds, while many learned checkpoints are trained once per
topology and architecture. This design makes the released comparison
auditable, but it does not replace multi-seed training, confidence
intervals over checkpoint selection, or architecture-specific
hyperparameter sweeps. Those additions would be necessary to make claims
about training stability rather than about the deployed checkpoints in
the benchmark artifact.

Finally, the benchmark deliberately compares \textbf{deployment roles}
that are not mathematically identical. Oracle is non-causal; MSSP-I uses
informed demand parameters; learned policies act through featurized
observations; heuristics encode structural assumptions; and LLM-Policy-C
uses low-frequency policy-parameter generation. This role asymmetry is
explicit rather than accidental, but it means the results should not be
read as if every controller solved the same information-relaxed problem.
The same caution applies to endogenous goodwill: for exogenous demand,
the Oracle is a perfect-foresight LP benchmark, whereas under goodwill
future demand depends on the policy-induced service path. In those rows,
the Oracle is a clairvoyant simulation-optimization benchmark rather than
a certified global upper bound.

\subsection{Open Challenges and Future Work}

These limitations point to extensions that build directly on the
released benchmark. At the evaluation layer, future releases should add
multi-seed training, bootstrap confidence intervals, paired
non-parametric tests, vector-valued rewards for multi-objective
trade-offs~\citep{felten2023toolkit}, and a JAX-native backend for
scalable rollouts~\citep{bonnet2024jumanji}; this release prioritizes
checkpoint-level auditability. The most immediate technical challenge is
\textbf{topology-aware action decoding}. The transfer experiments show that
graph encoders can preserve positive profit across unseen graphs, but
fixed-size action heads do not reliably route material to new retail
edges. Future graph policies should make action generation topology
native through per-edge decoders, pointer mechanisms~\citep{vinyals2015},
or edge-conditioned action heads.

A second direction is \textbf{hybrid OR--learning control}. This study
compares MSSP-I, DLP, heuristics, RL, residual RL, imitation, and LLM
policies as distinct paradigms; Residual RL is already a first integration
step because it learns a correction on top of a heuristic controller.
Future work can deepen the coupling by using stochastic programs to
produce scenario-aware base-stock or flow targets and by using learned
policies to warm-start or approximate rolling-horizon DLP/MSSP. Such
hybrids would combine OR quality references with amortized neural speed.

The benchmark also leaves room for \textbf{richer supply-chain physics and
data adapters}. Beyond the current DAGs, backlog/lost-sales regimes,
goodwill, fixed costs, and empirical traces, extensions should stress
heavy-tailed or mixture demand, longer 100--365 period horizons,
stochastic lead times, random yields, disruptions, shared resources,
liquidity limits, multi-product substitution, and price/calendar
covariates.

Finally, \textbf{foundation-model controllers} remain an open research
frontier rather than a solved benchmark component. The LLM experiments
suggest that raw per-period action generation is too slow and brittle,
whereas episode-level strategy extraction is more viable. Future work
should test constrained decoding, tool-verified JSON schemas, retrieval
over inventory histories, and hybrid LLM--OR policies where deterministic
controllers enforce feasibility and action bounds.

\subsection{Conclusion}

Taken together, the results argue for benchmark-driven method
selection. Informed stochastic programming is the strongest quality
reference when its information and compute requirements are acceptable;
heuristics remain highly competitive and transparent; learned policies
offer fast amortized decision making; residual and imitation methods
expose useful hybridization paths; and foundation models are currently
better suited to bounded strategy generation than direct control.
\texttt{gym-invmgmt} provides a common graph-based environment contract
for comparing these choices under identical simulator mechanics, making
the trade-offs visible rather than implicit.

\subsection*{CRediT Authorship Contribution Statement}

\textbf{Reza Barati:} Conceptualization, Methodology, Software,
Validation, Formal Analysis, Investigation, Data Curation, Writing
-- Original Draft, Visualization.
\textbf{Qinmin Vivian Hu:} Supervision, Writing -- Review \& Editing,
Resources, Project Administration.

\subsection*{Funding}

This research did not receive any specific grant from funding agencies
in the public, commercial, or not-for-profit sectors.

\subsection*{Declaration of Competing Interests}

The authors declare that they have no known competing financial
interests or personal relationships that could have appeared to
influence the work reported in this paper.

\subsection*{Code and Data Availability}

The benchmark source code, scenario definitions, evaluation scripts,
figure-generation utilities, and canonical merged result artifacts are
available at \url{https://github.com/r2barati/gym-invmgmt-paper}.
A standalone Gymnasium environment package is maintained at
\url{https://github.com/r2barati/gym-invmgmt}. Trained policy checkpoints
and matching normalization statistics are archived separately on Hugging
Face at \url{https://huggingface.co/datasets/rezabarati/gym-invmgmt-weights}.
The external retail datasets used for adapter validation are not
redistributed; scripts are provided to download them from their original
public sources.

\bibliographystyle{plainnat}
\bibliography{references}

\end{document}